\documentclass[11pt]{article}
 \RequirePackage{fix-cm}

\usepackage{amssymb, amsmath, amsfonts, mathtools}
\usepackage{epsfig}
\usepackage{graphicx}
\usepackage{subfigure}
\usepackage{epstopdf}
\usepackage{multicol}
\usepackage{array}
\usepackage{booktabs}

\usepackage{mathrsfs}
\usepackage {tikz}
\usetikzlibrary{patterns}
\usetikzlibrary {positioning}
\usetikzlibrary{arrows}
\usetikzlibrary{decorations.pathreplacing}
\usepackage {xcolor}

\newcolumntype{?}{!{\vrule width 1pt}}

\graphicspath{{images/}}

\begin{document}

\title{Modeling Uncertainty and Imprecision in Nonmonotonic Reasoning using Fuzzy Numbers 
}

\author{Sandip Paul\\
ECSU,Indian Statistical Institute, Kolkata
\and
Kumar Sankar Ray\\
ECSU,Indian Statistical Institute, Kolkata
\and
Diganta Saha\\
CSE Department\\
Jadavpur University}

\maketitle

\begin {abstract}

To deal with uncertainty in reasoning, interval-valued logic has been developed. But uniform intervals cannot capture difference in degrees of belief for different values in the interval. To salvage the problem triangular and trapezoidal fuzzy numbers are used as set of truth values along with traditional intervals. Preorder-based truth and knowledge ordering are defined over the set of fuzzy numbers defined over $[0,1]$. Based on this enhanced set of epistemic states, an answer set framework is developed, with properly defined logical connectives. This type of framework is efficient in knowledge representation and reasoning with vague and uncertain information under nonmonotonic environment where rules may posses exceptions.

\textbf{Keywords} Fuzzy numbers, Interval valued fuzzy sets, Preorder-based triangle, Answer Set Programming.

\end {abstract}

\section{Introduction}
Modern applications of artificial intelligence in decision support systems, plan generation systems require reasoning with imprecise and uncertain information. Logical frameworks based on bivalent reasoning are not suitable for such applications, because the set $\{0,1\}$ cannot capture the vagueness or uncertainty of underlying proposition. Though fuzzy logic-based systems can represent imprecise linguistic information by ascribing membership values to attributes (or truth values to propositions) taken from the interval [0,1], but this graded valuation becomes inadequate if the precise membership can not be determined due to some underlying uncertainty. This uncertainty may arise from lack of complete information or from lack of reliability of source of information or lack of unanimity among rational agents in a multi-agent reasoning system or from many other reasons. This uncertainty with respect to the assignment of membership degrees is captured by assigning a range of possible membership values, i.e. by assigning an interval. In other words by replacing the crisp \{0,1\} set by the set of sub-intervals of [0,1]. The intuition of such interval-valued system is that the actual degree, though still unknown, would be some value within the assigned interval and all the values in the interval are \textit{equally-likely} to be the actual one.

 However, there may be situations where all the values of an interval are not equally likely, rather, the information in hand suggests that some values are more plaussible. For instance, consider the motivating example presented in \cite{bauters2010towards}. It states that, "if the tumor suppressing genes(TSG) are lost due to mutation during cell division and chromosomal instability (CIN) is activated, a reasonably large tumor will grow". In the proposed approach, this single information is represented by four rules and the resultant valuation assigned to the fact \textit{tumor} is given by $\{tumor^{0.8:0.4}, tumor^{0.6:0.6}, tumor^{0.4:0.8}, tumor^{0.2:1}\}$.  This representation is very inefficient and the number of rules and number of elements in the valuation would grow proportionately to the number of truth degrees considered within $[0,1]$. This example denotes that in real-world applications assignment of uniform intervals is inadequate. Instead, if arbitrary distributions over the interval [0,1] are allowed for truth values of propositions that would hugely increase the expressibility of the system and reduce the number of necessary rules in the logic program. Therefore, instead of assigning a sub-interval of [0,1] as the epistemic state to some vague, uncertain proposition, a \textit{fuzzy number} defined over [0,1] would be a better choice, since, fuzzy numbers precisely allow to specify a membership distribution over [0,1]. 

Specifying the set of epistemic states is not enough, there has to be some underlying algebraic structure for ordering the values with respect to their degree of truth and degree of certainty(or uncertainty). For uniform interval-valued case Bilattice-based triangle structure were proposed \cite{cornelis2007uncertainty}. However later it is demonstrated \cite{ray2018preorder} that bilattice-based ordering is not suitable for belief revision in nonmonotonic reasoning and a preorder-based algebraic structure was constructed. Similar type of ordering has to be extended over the fuzzy numbers defined on $[0,1]$.

The main contributions of this work are as follows:

$\bullet$ The set of fuzzy numbers defined on [0,1] is considered as the set truth values for nonmonotonic reasoning with vague and uncertain information. In this work uniform, triangular and trapezoidal fuzzy numbers are considered only.

$\bullet$ Truth ordering and knowledge ordering over the set are defined (section 3) to construct the underlying preorder-based algebraic structure (section 4).

$\bullet$ This approach is used for answer set programming (section 5) to demonstrate the advantage.

\section{Fuzzy Numbers}

This section provides necessary preliminary concepts.

\textbf{Definition 1:}

A fuzzy set $A$ over some $X \subseteq \mathbb{R}$ is called a fuzzy number if

1. $A$ is convex, i.e.,

\begin{center}

$\mu_{A}(\lambda x_1 + (1-\lambda) x_2) \geq min(\mu_A(x_1), \mu_A(x_2))$

\end{center}

where, $x_1, x_2 \in X$ and $\lambda \in [0,1]$.

2. $A$ is normalised, i.e. $\max \mu_A(x) = 1$.

3. There is some $x \in X$ such that $\mu_A(x) = 1$.

4. $\mu_A(x)$ is piecewise continuous.
	
\subsection{Triangular and Trapezoidal Fuzzy Number}

The membership function of a triangular fuzzy number $TFN(a,b,c)$ for $a,b,c \in X $ and $a \leq b \leq c$ is specified as:

\begin{center}

$\mu_{(a,b,c)}(x) = \begin{cases} 0, & x < a \\ \frac{x-a}{b-a}, & x \in [a,b] \\ \frac{c-x}{c-b}, & x \in [b,c] \\ 0, & x > c \end{cases}$

\end{center}

The membership function of a trapezoidal fuzzy number $TrFN(a,b,c,d)$ for $a,b,c,d \in X $ and $a \leq b \leq c \leq d$ is specified as:

\begin{center}

$\mu_{(a,b,c,d)}(x) = \begin{cases} 0, & x < a \\ \frac{x-a}{b-a}, & x \in [a,b] \\ 1, & x \in [b,c] \\ \frac{d-x}{d-c}, & x \in [c,d] \\ 0, & x > d \end{cases}$

\end{center}

The uniform interval is a special case of $TrFN(a,b,c,d)$ when $a = b, c = d$, i.e., $TrFN(a,a,d,d)$ can be thought of as an interval $[a,d]$ so that all the values within the range has membership value 1. In this work an interval $[a,b]$ will be denoted as $IFN(a,b)$ to keep parity with the other two notations. 

\subsection{$\alpha$-cut decomposition of fuzzy numbers}

Another way of specifying a fuzzy number is by computing $\alpha$-cuts for $\alpha \in [0,1]$. For any fuzzy number $x$ and any specific value of $\alpha$, the $\alpha$-cut produces an interval of the form $x_{\alpha} = [\underline{x}_{\alpha}, \overline{x}_{\alpha}]$, where $\underline{x}_{\alpha}$ and $\overline{x}_{\alpha}$ are the intersection values with the left and right segment of $x$. The $\alpha$-cuts for a specific $\alpha$ for a TFN and TrFN are shown in Figure 1. $x_{\alpha}$ for $\alpha = 0$, will be referred to as \textit{base-range} of $x$ ($x_0$).

\begin{figure}

\begin{tikzpicture}

\draw[->] (0,0) -- (4,0);
\draw[->] (0,0) -- (0,3) node[left] {$\mu$};
\draw[-] (0.5,0) -- (1.5,2) -- (2.5,0);
\draw[dashed] (0,1.0) -- (2.5,1.0); 
\draw[dashed] (1.5,0) -- (1.5,2);
\node at (-0.2,0) {0};
\node at (-0.2,2) {1};
\node at (0.5,-0.2) {a};
\node at (1.5,-0.2) {b};
\node at (2.5,-0.2) {c};
\node at (-.2,1) {$\alpha$};
\node at (0.9, 1.2) {$\underline{x}_{\alpha}$};
\node at (2.2, 1.2) {$\overline{x}_{\alpha}$};
\node at (1.5,-1) {x = TFN(a,b,c)};

\begin{scope}[xshift=6.5cm]

\draw[->] (0,0) -- (4.5,0);
\draw[->] (0,0) -- (0,3) node[left] {$\mu$};
\draw[dashed] (0,1.0) -- (4.5,1.0); 
\draw[dashed] (1,0) -- (1,2);
\draw[dashed] (2.5,0) -- (2.5,2);
\node at (-0.2,0) {0};
\node at (-0.2,2) {1};
\node at (-.2,1) {$\alpha$};
\draw[-] (0.2,0) -- (1,2) -- (2.5,2) -- (4,0);
\node at (0.2,-0.2) {a};
\node at (1,-0.2) {b};
\node at (2.5,-0.2) {c};
\node at (4,-0.2) {d};
\node at (0.5, 1.2) {$\underline{y}_{\alpha}$};
\node at (3.5, 1.2) {$\overline{y}_{\alpha}$};

\node at (1.5,-1) {y = TrFN(a,b,c,d)};

\end{scope}

\end{tikzpicture}

\caption{Triangular and Trapezoidal Fuzzy Number with $\alpha$-cut}

\end{figure}
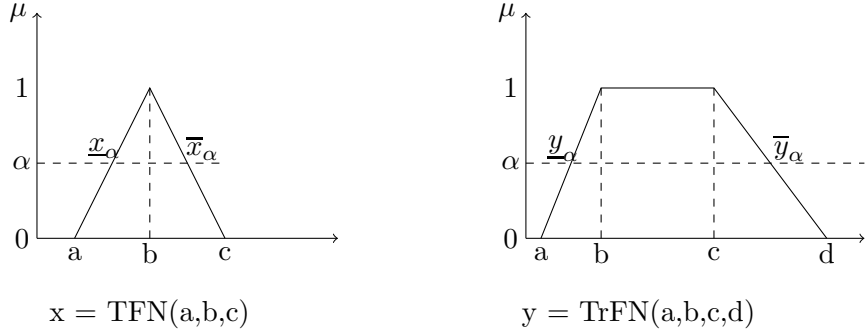

Analytically the $\alpha$-cut for the fuzzy numbers can be specified as follows:

$\bullet$ For $x = TFN(a,b,c)$; $x_{\alpha} = [a + \alpha(b-a), c - \alpha(c-b)]$;

$\bullet$ For $y = TrFN(a,b,c,d)$; $y_{\alpha} = [a + \alpha(b-a), d - \alpha(d-c)]$;

$\bullet$ For $z = IFN(a,d)$; $z_{\alpha} = [a,d]$.

Since, $IFN(a,d)$ is a special case of $TrFN(a,b,c,d)$ $z_{\alpha}$ can be obtained from $y_{\alpha}$ by setting $b = a$ and $c = d$. Similarly if the condition $b=c$ is imposed on $TrFN (a,b,c,d)$ a $TFN$ is obtained. Hence both are special cases of $TrFN$. Therefore, in later sections some concepts will be explained in terms of $TrFN$s only because same will be applicable for $IFN$ and $TFN$ by imposing the aforementioned conditions.

\section{Fuzzy numbers as truth assignment and their truth and knowledge ordering:}

It is already demonstrated by means of an example that specifying an interval of real numbers from [0,1] is not sufficient to express the epistemic state of propositions in real life reasoning with vague and uncertain information. Now, general fuzzy numbers can be used as truth assignment of a proposition to capture various degrees of belief over the range of $[0,1]$. However, just specifying fuzzy numbers as the set of epistemic states is not enough, there must be some ordering to order two such epistemic states with respect to the degree of truth (truth ordering) and degree of certainty (knowledge ordering). Instead of considering any general type of fuzzy numbers, here, only the three types, that are specified in Section 2 (i.e., IFN, TFN and TrFN), are considered as truth assignments.

\textbf{Definition 2:}

A $TrFN(a,b,c,d)$ is said to be \textit{restricted} if $0 \leq a,b,c,d \leq 1$, i.e., the base-range $x_0 \subseteq [0,1]$. . Similarly \textit{restricted} versions of $IFN$ and  $TFN$ are defined. 

A $TrFN(a,b,c,d)$ is \textit{semi-restricted} if $b,c \in [0,1]$ and $a < 0 \ \text{or} \ d > 1$ or both $a,d \notin [0,1]$. A $TFN(a,b,c)$ is \textit{semi-restricted} if $b \in [0,1]$ and any or both of $a$ and $c \notin [0,1]$.

\subsection{Construction of the Set of Epistemic States:}

In this section the set of truth assignments $\mathscr{T}$ is constructed so that any element from $\mathscr{T}$ can be assigned to some proposition to express its degree of belief. $\mathscr{T}$ is constructed from following conditions:

1. All restricted $TrFN$, $TFN$ and $IFN$ are member of $\mathscr{T}$.

2. For a semi-restricted $TrFN(a,b,c,d)$ its truncated version $[TrFN(a,b,c,d)]$ confined within [0,1] is included in $\mathscr{T}$.
Thus,

\begin{center}

$[TrFN(a,b,c,d)] = \begin{cases} TrFN(a,b,c,d), & x \in [0,1] \\ 0, & \text{otherwise} \end{cases}$

\end{center}

Here some intuitive aspects are explained to justify the necessity of $\mathscr{T}$ by means of examples.

\textbf{Example 1:} The case described in the introduction section can be re-considered. The epistemic state of the fact $tumor$ that was specified by $\{tumor^{0.8:0.4}, tumor^{0.6:0.6},\\ tumor^{0.4:0.8}, tumor^{0.2:1}\}$ can be approximately represented by $TFN(0.4,0.4,1.5)$. This assignment(shown in Figure 2a) is more compact representation.

\begin{figure}

\begin{tikzpicture}

\draw[->] (0,0) -- (4,0);
\draw[->] (0,0) -- (0,3) node[left] {$\mu$};
\draw[-] (0.8,0) -- (0.8,2) -- (2,0.9);
\draw[dashed] (2,0.9) -- (3,0);
\draw[-] (2,0) -- (2,0.9);
\draw[dashed] (2,0.9) -- (2,2.0); 
\node at (-0.2,0) {0};
\node at (-0.2,2) {1};
\node at (2,-0.2) {1};
\node at (0,-0.2) {0};

\draw[pattern=north west lines, pattern color=black] (0.8,0) -- (0.8,2) -- (2,0.9) -- (2,0);

\node at (1.5,-1) {(a) TFN(0.4,0.4,1.5)};

\begin{scope}[xshift=6.5cm]

\draw[->] (0,0) -- (3,0);
\draw[->] (0,0) -- (0,3) node[left] {$\mu$};
\node at (-0.2,0) {0};
\node at (-0.2,2) {1};

\draw[-] (0.8,0) -- (1.6,2) -- (2,2) -- (2,0);
\node at (0,-0.2) {0};
\node at (2,-0.2) {1};

\node at (1.5,-1) {(b) TrFN(0.4,0.8,1,1)};

\end{scope}

\end{tikzpicture}

\caption{Truth assignments for Example 1 and Example 2}

\end{figure}
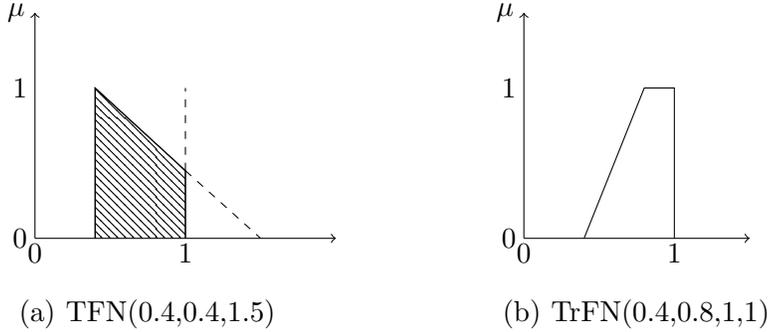

\textbf{Example 2:} Suppose a group of agents with different degree of expertise is asserting their degree of belief about some proposition $P$ under uncertainty. They all agree that $P$ is not false and has moderate to high degree of truth. The most reliable experts tend to ascribe very high degree of truth, which shows that they believe $P$ will be true. This scenario can be expressed by using a trapezoidal fuzzy number $TrFN(0.4,0.8,1,1)$, as shown in Figure 2b.

\textbf{Example 3:} If nothing is known about a proposition then $IFN(0,1)$ is assigned. If a proposition is known to be True, with absolute certainty, then $IFN(1,1)$ is assigned.

Bimodal or Multi-modal distributions can not be expressed using $\mathscr{T}$.

\subsection{Truth ordering and knowledge ordering of restricted $TrFN$s and restricted $TFN$s:} 

Now that the set of epistemic states $\mathscr{T}$ is specified and intuitively justified, elements of $\mathscr{T}$ are to be ordered with respect to their degree of truth and certainty. These orderings play crucial role in revising beliefs during nonmonotonic reasoning. For instance, suppose, based on available knowledge the truth status of certain proposition has been determined. Now, some additional information becomes available and based on the new enhanced information set, the proposition is re-evaluated. In such a scenario, it becomes important to compare the two new assignment with the previous one with respect to degree of truth and degree of certainty. If some contradiction arises, some previously known facts or rules are to be withdrawn and this withdrawal procedure mandates ordering various rules or facts with respect to their degree of certainty. It is demonstrated in \cite{ray2018preorder}, for $IFN$s preorder-based ordering is more intuitive and suitable for performing nonmonotonic reasoning with imprecise and uncertain information.

\textbf{Definition 3:}
For any two $IFN$, $[x_1,x_2]$ and $[y_1,y_2] \in \mathscr{T}$ the truth ordering($\leq_{t_p}$) and knowledge ordering($\leq_{k_p}$), defined in \cite{ray2018preorder}, are as follows:

\begin{center}
 $[x_1,x_2] \leq_{t_p}[y_1,y_2]\Leftrightarrow \frac{x_1+x_2}{2} \leq \frac{y_1+y_2}{2}$.

 $[x_1,x_2] \leq_{k_p}[y_1,y_2]\Leftrightarrow (x_2-x_1) \geq (y_2-y_1)$.
\end{center}

The truth ordering ($\leq_{t_p}$) and the knowledge ordering ($\leq_{k_p}$) are preorders and combined they give rise to a \textbf{preorder-based triangle}. These definitions are generalized for $TFN$s and $TrFN$s in the next subsections. 

\subsubsection{Truth-ordering}  

The intuition of assigning a fuzzy number for the epistemic state of a proposition, $p$ is that, due to uncertainty the actual truth assignment for $p$ (say, $\hat{p}$) is unknown, and hence is approximated by the assigned fuzzy number. If $\hat{p}$ is approximated by $IFN(a,b)$, then every value within the interval $[a,b]$ is equally probable to be $\hat{p}$. If $TFN(a,b,c)$ is assigned to $p$, then it signifies, in the range $[a,c]$, $b$ has a higher chance of being the actual truth status ($\hat{p}$) of $p$. Assignment of a $TrFN(a,b,c,d)$ can be interpreted similarly.

\newcolumntype{C}{>{\centering\arraybackslash}m{22em}}
\begin{table}\sffamily
\begin{tabular}{l*2{C}@{}}
\toprule
Assigned Fuzzy Number to $p$ & Equivalent probability density function of $\hat{p}$\\ 
\midrule

$IFN(a,b)$ &

$P_{IFN}(\hat{p}) = \begin{cases} \frac{1}{b-a}, & a \leq \hat{p} \leq b \\ 0, & \text{otherwise} \end{cases}$
\\

$TFN(a,b,c)$ &

$P_{TFN}(\hat{p}) = \begin{cases} 0, & \hat{p} < a \\ \frac{2(\hat{p}-a)}{(c-a)(b-a)}, & \hat{p} \in [a,b] \\ \frac{2(c-\hat{p})}{(c-1)(c-b)}, & \hat{p} \in [b,c] \\ 0, & \hat{p} > c \end{cases}$

\\

$TrFN(a,b,c,d)$ &

$P_{TrFN}(\hat{p}) = \begin{cases} 0, & \hat{p} < a \\ \frac{2(\hat{p}-a)}{(d+c-b-a)(b-a)}, & \hat{p} \in [a,b] \\ \frac{2}{d+c-b-a}, & \hat{p} \in [b,c] \\ \frac{2(d-\hat{p})}{(d+c-b-a)(d-c)}, & \hat{p} \in [c,d] \\ 0, & \hat{p} > d \end{cases}$ \\

\bottomrule 
\end{tabular}
\label{The Table}
\caption{Probability Distributions for Restricted Fuzzy Numbers in $\mathscr{T}$}
\end{table}

If we perform a random experiment, where an agent guesses the actual truth value of proposition $p$, then $\hat{p}$ can be thought of as a random variable, which follows a probability distribution. Now given the information in hand, assigning an epistemic state for $p$ is same as assigning an equivalent probability distribution over the random variable $\hat{p}$. So, for any restricted fuzzy number in $\mathscr{T}$, an equivalent probability distribution can be defined (as shown in Table 1).

For two propositions $p$ and $q$ with truth assignments $IFN(p_1,p_2)$ and $IFN(q_1,q_2)$ from $\mathscr{T}$, the intuition for ordering the truth assignments, with respect to the degree of truth is \cite{deschrijver2009generalized} 

\begin{center}

$IFN(p_1,p_2) \leq_{t_p} IFN(q_1,q_2)$ iff $Prob(\hat{p} \leq \hat{q}) \geq Prob(\hat{p} \geq \hat{q})$

\end{center}

where, $\hat{p}$ and $\hat{q}$ stands for the actual (yet unknown) truth status of propositions $p$ and $q$ respectively.

Now following this intuition we intend to extend the truth ordering from Definition 3 to ordering $TFN$s and $TrFN$s.

As explained above, for two propositions $p$ and $q$, $\hat{p}$ and $\hat{q}$ can be thought of as two random variables. In order to calculate $Prob(\hat{p} \leq \hat{q})$ or $Prob(\hat{p} \geq \hat{q})$ another random variable $\hat{r}$ is defined as:

\begin{center}

$\hat{r} = \hat{p} - \hat{q}$.

\end{center}

Then, $Prob(\hat{p} \leq \hat{q}) = Prob(\hat{r} \leq 0)$ and $Prob(\hat{p} \geq \hat{q}) = Prob(\hat{r} \geq 0)$. Moreover the expectations(or means) of the random variables are related by $E(\hat{r}) = E(\hat{p}) - E(\hat{q})$. 

Now, if probability distributions of $\hat{p}$ and $\hat{q}$ are chosen so that $E(\hat{p}) = E(\hat{q})$, then $E(\hat{r}) = 0$. This makes, $Prob(\hat{r} \leq 0) = Prob(\hat{r} \leq E(\hat{r})) = Prob(\hat{r} \geq E(\hat{r})) = Prob(\hat{r} \geq 0)$. Thus, $Prob(\hat{p} \leq \hat{q}) = Prob(\hat{p} \geq \hat{q})$. Since, the truth ordering is a total preorder, this would signify that propositions $p$ and $q$ have \textit{same} degree of truth. This occurs irrespective of the chosen probability distribution of $\hat{p}$ and $\hat{q}$.

\textbf{Definition 4:} For any member $\mathscr{P} \in \mathscr{T}$, its \textbf{equivalent-interval}$(Eq-int)$ is any restricted $IFN(a,b)$ so that mean value of the equivalent probability distribution of $\mathscr{P}$ is equal to $\frac{a+b}{2}$, i.e., the expected value of a random variable $X$ that follows the probability density function $P_{IFN(a,b)}$. Therefore, any $IFN \in \mathscr{T}$ centered around the value $\frac{a+b}{2}$ is an equivalent-interval to $\mathscr{P}$.

The truth ordering defined over $IFN$s (from Definition 3) can be extended for ordering restricted $TrFn$s and $TFN$s using their equivalent-intervals.

\textbf{Theorem 1:} For any members $\mathscr{P}_1, \mathscr{P}_2 \in \mathscr{T}$,

\begin{center}

$\mathscr{P}_1 \leq_{t_p} \mathscr{P}_2$ iff $E(X_{\mathscr{P}_1}) \leq E(X_{\mathscr{P}_2})$.

\end{center}

where, $X_{\mathscr{P}_1}$ and $X_{\mathscr{P}_2}$ are random variables following probability density functions equivalent to $\mathscr{P}_1$ and $\mathscr{P}_2$(as specified in Table 1) respectively.

\textbf{Proof:} If $\mathscr{P}_1$ and $\mathscr{P}_2$ are $IFN$s then the theorem directly follows from Definition 3, as $E(X_{IFN(a,b)}) = \frac{a+b}{2}$.

Suppose, $\mathscr{P}_1$ and $\mathscr{P}_2$ are respectively $Trapz_1 = TrFN(a_1,b_1,c_1,d_1)$ and $Trapz_2 = TrFN(a_2,b_2,c_2,d_2)$, and their corresponding equivalent-intervals are $Eq-int_1$ and $Eq-int_2$ respectively. Following the aforementioned rationale $Trapz_1$ and $Eq-int_1$ have same degree of truth and same holds for $Trapz_2$ and $Eq-int_2$. The two $IFN$s, $Eq-int_1$ and $Eq-int_2$ can be ordered with respect to $(\leq_{t_p})$ following Definiton 3.Thus,

\begin{center}

$TrFN(a_1,b_1,c_1,d_1) \leq_{t_P} TrFN(a_2,b_2,c_2,d_2)$ iff $Eq-int_1 \leq_{t_p} Eq-int_2$.
 
\end{center}

In other words,

$Trapz_1 \leq_{t_P} Trapz_2$ iff $E(X_{Eq-int_1}) \leq E(X_{Eq-int_2})$,

$\Rightarrow Trapz_1 \leq_{t_P} Trapz_2$ iff $E(X_{Trapz_1}) \leq E(X_{Trapz_2})$

Since, following Definition 4, $E(X_{Trapz_1}) = E(X_{Eq-int_1})$ and $E(X_{Trapz_2}) = E(X_{Eq-int_2})$.

$TFN$s being special cases of $TrFN$s the theorem can similarly be proved if $\mathscr{P}_1$ and $\mathscr{P}_2$ are $TFN$s, or if $\mathscr{P}_1$ is an $IFN$ and $\mathscr{P}_2$ is a $TFN$ or a $TrFN$ as well. (\textbf{Q.E.D})

Theorem 1 essentially gives the definition of preorder-based truth ordering ($\leq_{t_p}$) of restricted fuzzy numbers of $\mathscr{T}$. Therefore, for any restricted fuzzy numbers $\mathscr{P}_1, \mathscr{P}_2 \in \mathscr{T}$;

$\bullet$ $\mathscr{P}_1 = IFN(a_1,b_1), \mathscr{P}_2 = IFN(a_2,b_2)$;

\begin{center}

$\mathscr{P}_1 \leq_{t_p} \mathscr{P}_2$ iff $\frac{a_1+b_1}{2} \leq \frac{a_2+b_2}{2}$.

\end{center}

$\bullet$ $\mathscr{P}_1 = TFN(a_1,b_1,c_1), \mathscr{P}_2 = TFN(a_2,b_2,c_2)$;

\begin{center}

$\mathscr{P}_1 \leq_{t_p} \mathscr{P}_2$ iff $\frac{a_1+b_1+c_1}{3} \leq \frac{a_2+b_2+c_2}{3}$.

\end{center}

$\bullet$ $\mathscr{P}_1 = TrFN(a_1,b_1,c_1,d_1), \mathscr{P}_2 = TrFN(a_2,b_2,c_2,d_2)$;

\begin{center}

$\mathscr{P}_1 \leq_{t_p} \mathscr{P}_2$ iff $\frac{1}{3(d_1+c_1-b_1-a_1)}(\frac{d_1^3-c_1^3}{d_1-c_1} - \frac{b_1^3-a_1^3}{b_1-a_1}) \leq \frac{1}{3(d_2+c_2-b_2-a_2)}(\frac{d_2^3-c_2^3}{d_2-c_2} - \frac{b_2^3-a_2^3}{b_2-a_2})$.

\end{center}

$\bullet$ $\mathscr{P}_1 = TFN(a_1,b_1,c_1), \mathscr{P}_2 = TrFN(a_2,b_2,c_2,d_2)$;

\begin{center}

$\mathscr{P}_1 \leq_{t_p} \mathscr{P}_2$ iff $\frac{a_1+b_1+c_1}{3} \leq \frac{1}{3(d_2+c_2-b_2-a_2)}(\frac{d_2^3-c_2^3}{d_2-c_2} - \frac{b_2^3-a_2^3}{b_2-a_2})$.

\end{center}

\textbf{Example 4:} This example analytically validates Theorem 1. Consider two truth assignments $\mathscr{P} = IFN(a,d)$ and $\mathscr{Q} = TrFN(a,b,c,d)$, with $\hat{p}$ and $\hat{q}$ being their actual truth values approximated by $\mathscr{P}$, $\mathscr{Q}$ respectively. The actual truth status $\hat{p}$ and $\hat{q}$ are independent random variables, that follow a uniform and a trapezoidal probability density functions $P_{IFN(a,b)}$ and $P_{TrFN(a,b,c,d)}$ respectively. So, $P_{\mathscr{P}} = P_{IFN(a,d)}$ and $P_{\mathscr{Q}} = P_{TrFN(a,b,c,d)}$. The joint probability density function $f_{\mathscr{P}\mathscr{Q}} = P_{\mathscr{P}}P_{\mathscr{Q}}$.

$Prob(\hat{p} \leq \hat{q}) = \int_{a}^{d} \int_{\hat{p}}^{d} f_{\mathscr{P}\mathscr{Q}}(\hat{p},\hat{q}) d\hat{q}d\hat{p}$

$=\int_{a}^{d} \int_{\hat{p}}^{d} P_{\mathscr{P}}(\hat{p}).P_{\mathscr{Q}}(\hat{q}) d\hat{p}d\hat{q}$,

$=\frac{1}{d-a} \int_{a}^{d} \int_{\hat{p}}^{d} P_{\mathscr{Q}}(\hat{q}) d\hat{q}d\hat{p}$,

$= \frac{1}{d-a} \int_{a}^{b} \int_{\hat{p}}^{d} P_{\mathscr{Q}}(\hat{q})d\hat{q}d\hat{p} + \frac{1}{d-a} \int_{b}^{c} \int_{\hat{p}}^{d} P_{\mathscr{Q}}(\hat{q})d\hat{q}d\hat{p} + \frac{1}{d-a} \int_{c}^{d} \int_{\hat{p}}^{d} P_{\mathscr{Q}}(\hat{q})d\hat{q}d\hat{p}$,

$= \frac{1}{d-a} \int_{a}^{b}[1-\frac{(\hat{p}-a)^2}{(d-a+c-b)(b-a)}]d\hat{p} + \frac{1}{d-a} \int_{b}^{c}[\frac{2(c-\hat{p})}{d-a+c-b} + \frac{d-c}{d-a+c-b}]d\hat{p} + \frac{1}{d-a} \int_{c}^{d} \frac{d-\hat{p}}{(d-a+c-b)(d-c)}d\hat{p}$,

$=\frac{3(d-a+c-b)(b-a) - (b-a)^2}{3(d-a+c-b)(d-a)} + \frac{(c-b)(d-b)}{(d-a+c-b)(d-a)} + \frac{(d-c)^2}{3(d-a+c-b)(d-a)}$,

$= \frac{-ab-b^2+cd-3ad+2a^2-3ac+3ab+d^2+c^2}{3(d-a+c-b)(d-a)}$.

Now, $Prob(\hat{p} \leq \hat{q}) \geq Prob(\hat{p} \geq \hat{q})$

$\Rightarrow Prob(\hat{p} \leq \hat{q}) \geq \frac{1}{2}$,

$\Rightarrow \frac{-ab-b^2+cd-3ad+2a^2-3ac+3ab+d^2+c^2}{3(d-a+c-b)(d-a)} \geq \frac{1}{2}$

$\Rightarrow a^2-2b^2+2c^2-d^2+ad-3ac-cd+3bd \geq 0$

$\Rightarrow 2d^2 + 2cd + 2c^2 - 2b^2 - 2ab - 2a^2 \geq 3ac - 3a^2 - 3ab + 3d^2 + 3cd - 3bd$,

$\Rightarrow 2(\frac{d^3-c^3}{d-c} - \frac{b^3 - a^3}{b-a}) \geq 3a(d+c-b-a) + 3d(d-a+c-b)$,

$\Rightarrow \frac{1}{3(d+c-b-a)}(\frac{d^3-c^3}{d-c} - \frac{b^3 - a^3}{b-a}) \geq \frac{a+d}{2}$,

$\Rightarrow E(\hat{q}) \geq E(\hat{p}).$

As a special case, having $b=c$ in $TrFN(a,b,c,d)$ gives $\mathscr{Q} = TFN(a,b,d)$. Putting this condition in the above derivation gives,

$Prob(\hat{p} \leq \hat{q}) \geq \frac{1}{2}$

$\Rightarrow a^2 - d^2 - 2ab + 2bd \geq 0$

$\Rightarrow 2b(d-a) - (a+d)(d-a) \geq 0$

$\Rightarrow b \geq \frac{a+d}{2}$

$\Rightarrow a+b+d \geq \frac{3(a+d)}{2}$

$\Rightarrow \frac{a+b+d}{3} \geq \frac{a+d}{2}$

$\Rightarrow E(\hat{q}) \geq E(\hat{p})$.

Consider three propositions $p$, $q_1$ and $q_2$, ascribed with $\mathscr{P} = IFN(0.3,0.7)$, $\mathscr{Q}_1 = TrFN(0.3,0.3,0.5,0.7)$ and $\mathscr{Q}_2 = TrFN(0.3,0.5,0.7,0.7)$.

$E(\hat{p}) = 0.5$, $E(\hat{q}_1) = 0.455$, $E(\hat{q}_2) = 0.56$. It can be seen, $E(\hat{p}) \geq E(\hat{q}_1)$ and $E(\hat{p}) \leq E(\hat{q}_2)$. Also, $Prob(\hat{p} \leq \hat{q}_2) = 0.617 > 0.5$ and $Prob(\hat{p} \leq \hat{q}_1) = 0.388 < 0.5$.

So, $\mathscr{P} \leq_{t_p} \mathscr{Q}_2$ and $\mathscr{Q}_1 \leq_{t_p} \mathscr{P}$.

\subsubsection{Knowledge-ordering}

As evident from Definition 3, the knowledge ordering is based on the length of $IFN$s, i.e., more is the length more is the underlying uncertainty. Therefore, the length of an $IFN$ identifies its level of uncertainty. 

\textbf{Uncertainty degree of TFN:}

The concept of length is not so obvious for $TFN$ as it is for $IFN$s. To do so, the $\alpha$-cut decomposition of $TFN$ is used. 

For $x = TFN(a,b,c)$ the $\alpha$-cut for any any value of $\alpha$ is an $IFN$ given as $x_{\alpha} = [\underline{x}_{\alpha}, \overline{x}_{\alpha}] = [a + \alpha(b-a), c - \alpha(c-b)]$. Now $x_{\alpha}$ being an $IFN$, the degree of its uncertainty can be evaluated to be:

\begin{center}
$k_{x_{\alpha}} = [\overline{x}_{\alpha} - \underline{x}] = (c-a) - \alpha(c-b+b-a) = (c-a)-\alpha(c-a)$. 
\end{center}

$k_{x_{\alpha}}$ varies with different values of $\alpha$ in [0,1]. Hence the average uncertainty(or length) is obtained as:

\begin{center}

$k_x = \int_{0}^{1}[(c-a)-\alpha(c-a)] d\alpha = \frac{(c-a)}{2}$

\end{center}

Thus for two $TFN$s in $\mathscr{T}$, namely $\mathscr{P}_1 = TFN(a_1,b_1,c_1)$ and $\mathscr{P}_2 = TFN(a_2,b_2,c_2)$; it can be said,

\begin{center}

$TFN(a_1,b_1,c_1) \leq_{k_p} TFN(a_2,b_2,c_2) \Leftrightarrow \frac{(c_1-a_1)}{2} \geq \frac{(c_2-a_2)}{2}$

\end{center}

\textbf{Uncertainty degree of TrFN:}

For $y = TrFN(a,b,c,d)$ and for some $\alpha$ in $\in [0,1]$; $y_{\alpha} = [a + \alpha(b-a), d - \alpha(d-c)]$ and $k_{y_{\alpha}} = (d-a) - \alpha(d-c+b-a)$. Therefore,

\begin{center}

$k_y = \int_{0}^{1}[(d-a)-\alpha(d-c+b-a)] d\alpha = \frac{d+c-b-a}{2}$.

\end{center}

Hence, for $TrFN(a_1,b_1,c_1,d_1)$ and $TrFN(a_2.b_2,c_2,d_2)$,

\begin{center}

$TrFN(a_1,b_1,c_1,d_1) \leq_{k_p} TrFN(a_2,b_2,c_2,d_2) \Leftrightarrow \frac{(d_1+c_1-b_1-a_1)}{2} \geq \frac{(d_2+c_2-b_2-a_2)}{2}$.

\end{center}

In a nutshell, 

$\bullet$ uncertainty degree of $IFN(a,b)$, $k_{IFN} = b-a$;

$\bullet$ uncertainty degree of $TFN(a,b,c)$, $k_{TFN} = \frac{c-a}{2}$;

$\bullet$ uncertainty degree of $TFN(a,b,c,d)$, $k_{TrFN} = \frac{d+c-b-a}{2}$.

For any restricted fuzzy numbers $\mathscr{P}_1, \mathscr{P}_2 \in \mathscr{T}$,

\begin{center}

$\mathscr{P}_1 \leq_{k_p} \mathscr{P}_2$ iff $k_{\mathscr{P}_1} \geq k_{\mathscr{P}_2}$.

\end{center}

\textbf{Example 5:} Consider $\mathscr{P}_1$, $\mathscr{P}_2$, $\mathscr{P}_3 \in \mathscr{T}$ and $\mathscr{P}_1 = IFN(a,d)$, $\mathscr{P}_2 = TFN(a,b,d)$ and $\mathscr{P}_3 = TrFN(a,c,e,d)$.

Now, $d-a \geq \frac{d+e-c-a}{2} \geq \frac{d-a}{2}$, i.e., $k_{\mathscr{P}_1} \geq k_{\mathscr{P}_3} \geq k_{\mathscr{P}_2}$. Therefore,

\begin{center}

$\mathscr{P}_1 \leq_{k_p} \mathscr{P}_3 \leq_{k_p} \mathscr{P}_3$.

\end{center}

This is intuitive, since in case of $IFN(a,d)$ all values in $[a,d]$ are equally probable, whereas for $TFN(a,b,d)$, $b$ is more likely than any other value in $[a,d]$; which means $TFN(a,b,d)$ provides more information about the truth status of the underlying proposition than $IFN(a,d)$. $TrFN(a,c,e,d)$ lies in between.

\textbf{Note}: One notable point is that the uncertainty degrees of a fuzzy number as calculated is actually equal to the underlying area of the membership function of the fuzzy number. When there is no uncertainty and a specific membership value is assigned then the uncertainty degree is zero and so is the area under the curve of the form $IFN(a,a)$, for some $a \in [0,1]$. This can be utilised for calculating the uncertainty degree of semi-restricted $TFN$s and $TrFN$s.

\subsection{Truth ordering and knowledge ordering of truncated semi-restricted $TrFN$s and $TFN$s in $\mathscr{T}$:}

The notion of truth and knowledge ordering, as defined over restricted fuzzy numbers of $\mathscr{T}$, can be extended to every pair of members of $\mathscr{T}$.

\subsubsection{Uncertainty degree and knowledge ordering:}

For semi-restricted fuzzy numbers their truncated versions within the interval [0,1] are considered. Therefore, the expressions for uncertainty degree as specified in section 3.2.2 is no longer valid if the base-range of the fuzzy number exceeds [0,1]. However, from the notion developed in previous subsection, the uncertainty degree can be easily calculated from evaluating the area underlying the curve in the interval [0,1]. The more general expressions for uncertainty degree of elements of $\mathscr{T}$ are presented here.

For $[TFN(a,b,c)]$,

\begin{center}
 
$k_{[TFN]} = \begin{cases} \frac{c-a}{2} - \frac{a^2}{2(b-a)}, & a < 0, c \in [0,1] \\ \frac{c-a}{2} - \frac{(c-1)^2}{2(c-b)}, & a \in [0,1], c > 1 \\ \frac{c-a}{2} - \frac{a^2}{2(b-a)} - \frac{(c-1)^2}{2(c-b)}, & a < 0, c > 1 \end{cases}$

\end{center}

For $[TrFN(a,b,c,d)]$,

\begin{center}

$k_{[TrFN]} = \begin{cases} \frac{d+c-a-b}{2} - \frac{a^2}{2(b-a)}, & a < 0, d \in [0,1] \\ \frac{d+c-a-b}{2} - \frac{(d-1)^2}{2(d-c)}, & a \in [0,1], d > 1 \\ \frac{d+c-a-b}{2} - \frac{a^2}{2(b-a)} - \frac{(d-1)^2}{2(d-c)}, & a < 0, d > 1 \end{cases}$

\end{center}

In general, for any $TFN$ in $\mathscr{T}$, the knowledge degree can be specified as:

\begin{center}

$k_{\Delta} = \frac{c-a}{2} - \frac{(min(a,0))^2}{2(b-a)} - \frac{(max(c,1)-1)^2}{2(c-b)}$.

\end{center}

For any $TrFN \in \mathscr{T}$;

\begin{center}

$k_{\Box} = \frac{d+c-b-a}{2} - \frac{(min(a,0))^2}{2(b-a)} - \frac{(max(d,1)-1)^2}{2(d-c)}$

\end{center}

Based on the uncertainty degree the knowledge ordering can be induced in the same way as mentioned in the previous subsection.

\textbf{Note:} $\Delta$ and $\Box$ notations are used to denote both restricted and truncated semi-restricted triangular or trapezoidal fuzzy numbers respectively in general.

\subsubsection{Equivalent probability distribution and Truth ordering:}

Suppose a fuzzy number $\mathscr{P}_g$ (which may be restricted or truncated semi-restricted) is assigned as epistemic state to some proposition $p$, then the underlying probability density function of the actual and unknown truth degree $\hat{p}$ can be defined as follows:

$\bullet$ If $\mathscr{P}_g = \Delta(a,b,c)$, then

\begin{center}

$P_{\Delta(a,b,c)}(\hat{p}) = \begin{cases} h_{\Delta}\frac{\hat{p}-a}{b-a}, & max(0,a) \leq \hat{p} \leq b \\ h_{\Delta}\frac{c-\hat{p}}{c-b}, & b \leq \hat{p} \leq min(c,1) \\ 0, & \text{otherwise} \end{cases}$

\end{center}

where, $h_{\Delta} = \frac{1}{k_{\Delta}}$.

$\bullet$ If $\mathscr{P}_g = \Box(a,b,c,d)$, then

\begin{center}

$P_{\Box(a,b,c,d)}(\hat{p}) = \begin{cases} h_{\Box}\frac{\hat{p}-a}{b-a}, & max(0,a) \leq \hat{p} \leq b \\ h_{\Box}, & b \leq \hat{p} \leq c \\ h_{\Box}\frac{d-\hat{p}}{d-c}, & c \leq \hat{p} \leq min(d,1) \\ 0, & \text{otherwise} \end{cases}$

\end{center}

where, $h_{\Box} = \frac{1}{k_{\Box}}$.

For specifying the truth-ordering Theorem 1 is used. The expected values for random variables following the above-mentioned probability density functions can be calculated as:

$\bullet$ If $\hat{p}$ follows $P_{\Delta(a,b,c)}$, then

\begin{center}

$E(\hat{p}) = \int_{max(0,a)}^{b} \frac{\hat{p}(\hat{p}-a)h_{\Delta}}{(b-a)}d\hat{p} + \int_{b}^{min(c,1)} \frac{\hat{p}(c-\hat{p})h_{\Delta}}{(c-b)} d \hat{p}$.

\end{center} 

$\bullet$ If $\hat{p}$ follows the pdf $P_{\Box(a,b,c,d)}$, then

\begin{center}

$E(\hat{p}) = \int_{max(a,0)}^b \frac{h_{\Box}}{(\hat{p}-a)\hat{p}}{(b-a)} d\hat{p} + \int_{b}^{c} h_{\Box} d\hat{p} + \int_{c}^{min(d,1)} \frac{\hat{p}(d-\hat{p})h_{\Box}}{d-c} d\hat{p}$.

\end{center}

Once the mean value is calculated the truth ordering can be induced based on Theorem 1.

\textbf{Example 6:} The fuzzy number $[TFN](0.4,0.4,1.5)$ shown in Figure 2(a) is a semi-restricted element of $\mathscr{T}$, assigned to a proposition, say $p$. 

Now, the uncertainty degree $k_{[TFN]} = \frac{(1.5-.4)}{2} - \frac{(1.5-1)^2}{2(1.5-0.4)} = 0.436$; thus $h_{[TFN]} = \frac{1}{0.436} = 2.292$. 

Truth degree  = $E(\hat{p}) = $

\begin{center}

$\int_{0.4}^{1} 2.083\hat{p}(1.5 - \hat{p}) d\hat{p} = 0.574$.

\end{center}

\section{Preorder-based Triangle for $\mathscr{T}$:}

Each element $\mathscr{P} \in \mathscr{T}$ can be seen as a pair $(t_{\mathscr{P}}, k_{\mathscr{P}})$, where, $t_{\mathscr{P}}$ is the truth degree of $\mathscr{P}$, which is the expected value (mean) of a random variable following the probability density function $P_{\mathscr{P}}$ and $k_{\mathscr{P}}$ is the uncertainty degree.

For any two members $\mathscr{P}_1, \mathscr{P}_2 \in \mathscr{T}$:

$\mathscr{P}_1 \leq_{t_p} \mathscr{P}_2  \ \text{iff}  \ t_{\mathscr{P}_1} \leq t_{\mathscr{P}_2}$

$\mathscr{P}_1 \leq_{k_p} \mathscr{P}_2  \ \text{iff}  \ k_{\mathscr{P}_1} \geq k_{\mathscr{P}_2}$

These two orderings imposed on the elements of $\mathscr{T}$ give rise to a \textit{Preorder-based triangle} for $\mathscr{T}$, \textbf{P}($\mathscr{T}$), which can be thought as an extension of the structure developed in \cite{ray2018preorder}. 

\textbf{Example 7:} Consider a lattice \textbf{L}$ = [\{0,\frac{1}{3}, \frac{2}{3}, 1\}, \leq]$. Let $\mathscr{T}_R(\textbf{L})$ be the set of $IFN$s and restricted $TFN$s and $TrFN$s constructed from \textbf{L}. The elements of $\mathscr{T}_R(L)$ are shown in table

\begin{table*}[h!]
\begin{center}	
\begin{tabular}{|c|c?c|c?c|c|}

\hline 

\textbf{IFN} & $(t_{I},k_{I})$ & \textbf{TFN} & $(t_{\Delta},k_{\Delta})$ & \textbf{TrFN} & $(t_{\Box},k_{\Box})$\\

\hline

1. [0,0] & (0,0) & 1.(0,$\frac{1}{3}$,1) & ($\frac{4}{9}$,$\frac{1}{2}$) & 1.(0,$\frac{1}{3}$,$\frac{2}{3}$,1) & ($\frac{1}{2}$, $\frac{2}{3}$)\\

2. [$\frac{1}{3}$, $\frac{1}{3}$] & ($\frac{1}{3}$,0) & 2. (0,$\frac{1}{3}$,$\frac{2}{3}$) & ($\frac{1}{3}$,$\frac{1}{3}$) & 2. (0,0,$\frac{1}{3}$,$\frac{2}{3}$) & ($\frac{7}{27}$, $\frac{1}{2}$)\\

3. [$\frac{2}{3}$, $\frac{2}{3}$] & ($\frac{2}{3}$,0) & 3. (0,$\frac{2}{3}$,1) & ($\frac{5}{9}$,$\frac{1}{2}$) & 3. (0,0,$\frac{2}{3}$,1) & ($\frac{19}{45}$, $\frac{5}{6}$)\\

4. [1,1] & (1,0) & 4. (0,0,$\frac{2}{3}$) & ($\frac{2}{9}$,$\frac{1}{3}$) & 4. ($\frac{1}{3}$,$\frac{1}{3}$,$\frac{2}{3}$,1) & ($\frac{16}{27}$, $\frac{1}{2}$)\\

5. [0,$\frac{1}{3}$] & ($\frac{1}{6}$,$\frac{1}{3}$) & 5. (0,0,1) & ($\frac{1}{3}$,$\frac{1}{2}$) & 5. (0,$\frac{1}{3}$,1,1) & ($\frac{26}{45}$, $\frac{5}{6}$)\\

6. [0,$\frac{2}{3}$] & ($\frac{1}{3}$,$\frac{2}{3}$) & 6. ($\frac{1}{3}$,1,1) & ($\frac{7}{9}$,$\frac{1}{3}$) & 6. (0,$\frac{2}{3}$,1,1) & ($\frac{23}{36}$, $\frac{2}{3}$)\\

7. [0,1] & ($\frac{1}{2}$,1) & 7. (0,1,1) & ($\frac{2}{3}$,$\frac{1}{2}$) & 7. ($\frac{1}{3}$,$\frac{2}{3}$,1,1) & ($\frac{20}{27}$, $\frac{1}{2}$)\\

8. [$\frac{1}{3}$,$\frac{2}{3}$] & ($\frac{1}{2}$,$\frac{1}{3}$) & 8. ($\frac{1}{3}$,$\frac{2}{3}$,1) & ($\frac{2}{3}$,$\frac{1}{3}$) & 8. (0,$\frac{1}{3}$,$\frac{2}{3}$,$\frac{2}{3}$) & ($\frac{11}{27}$, $\frac{1}{2}$)\\

9. [$\frac{1}{3}$,1] & ($\frac{2}{3}$,$\frac{2}{3}$) & 9. ($\frac{1}{3}$,$\frac{1}{3}$,1) & ($\frac{5}{9}$,$\frac{1}{3}$) & 9. (0,0,$\frac{1}{3}$,1) & ($\frac{13}{36}$, $\frac{2}{3}$)\\

10. [$\frac{2}{3}$,1] & ($\frac{5}{6}$,$\frac{1}{3}$) & 10. (0,$\frac{2}{3}$,$\frac{2}{3}$) & ($\frac{4}{9}$,$\frac{1}{3}$) & & \\

\hline

	\end{tabular}
	\end{center}
	\caption{Elements of $\mathscr{T}_R(L)$ and their truth and uncertainty degrees}
\end{table*}

\begin{figure}
\begin{center}
\includegraphics [width=100mm]{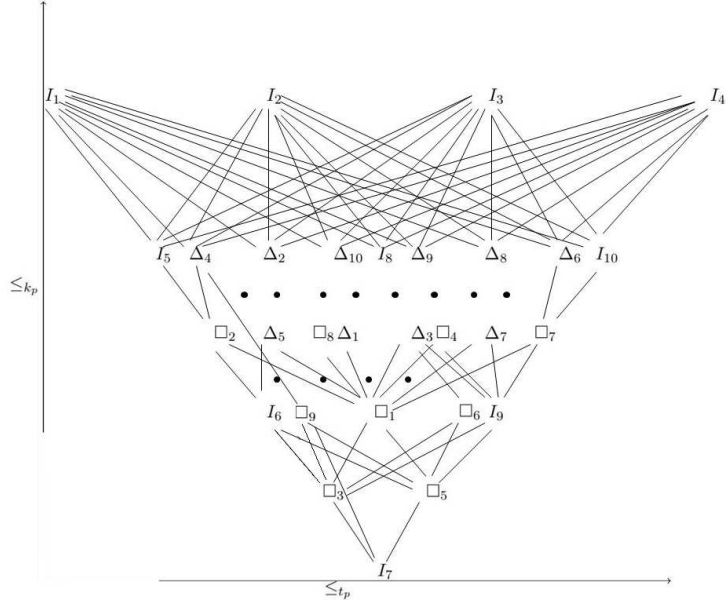}
\caption{Preorder-based Triangle for $\mathscr{T}_R(L)$}
\label{fig:tri}
\end{center}
\end{figure}

Figure 3 shows the preorder-based triangle for $\mathscr{T}_R(L)$ constructed from the truth ordering($\leq_{t_p}$) and knowledge ordering($\leq_{k_p}$) as specified. However, for clarity, all the comparibility connections are not shown in the figure. If truncated semi restricted fuzzy numbers were included in $\mathscr{T}_R(L)$ it would not change the "triangular" nature of the algebraic structure, rather that would increase the number of available epistemic states in $\mathscr{T}_R(L)$. 

Evidently, the extension of set of truth values from just the set of sub-intervals of $[0,1]$ to $\mathscr{T}$ introduces many new epistemic states.

\section{Answer set programming using $\mathscr{T}$ as set of truth values:}

Answer Set Programming (ASP) \cite{lifschitz2008answer,baral2003knowledge} is a nonmonotonic reasoning framework which is hugely used in declarative problem solving and reasoning with rules having exceptions. Numerous Fuzzy \cite{janssen2009general,janssen2012core,blondeel2014complexity,mushthofa2014finite} and possibilistic \cite{bauters2012possibilistic,bauters2015characterizing} extensions of ASP are developed for dealing with real-life problems that encounter imprecise and uncertain information. 

In \cite{bauters2010towards} a possibilistic fuzzy ASP framework is proposed. However, as mentioned earlier, the knowledge representation of the proposed framework is inefficient. Moreover, the approach is developed only for programs with positive rules and no negation has been introduced in the system. A more concrete and intuitive approach, namely Unified Answer set Programming has been reported \cite{paul2019unified}, that uses interval-valued fuzzy sets (IVFS) defined over the unit interval $[0,1]$, as the set of truth values. Replacing IVFS with $\mathscr{T}$, as developed in the previous section, would enrich the Unified Answer Set Programming framework in terms of its expressing ability and intuitive knowledge representation.

Answer set programming with $\mathscr{T}$ (defined over some lattice \textbf{L}) as the truth value space is described in this section.

\subsection{Logical Operators:}

The logical operators are defined based on the traditional operations defined over [0,1] in Fuzzy logic and the algebraic operations on fuzzy numbers. Traditional fuzzy t-(co)norms, negation and the operators defined in \cite{paul2019unified} can be obtained as special cases of the operators defined here.

\subsubsection{Negation:}

For a $\mathscr{P} = \Box(a,b,c,d) \in \mathscr{T}$, its negation is defined as:

\begin{center}

$\neg \mathscr{P} = \neg \Box(a,b,c,d) = \Box(1-d,1-c,1-b,1-a)$.

\end{center}

This negation is involutive. This is the generalised version of the standard negator defined over IVFS \cite{cornelis2007uncertainty}, which is $\neg[a,b] = [1-b,1-a]$. Negation of a certain assertion, denoted by an exact interval $IFN(x,x)$ for $x \in [0,1]$ becomes $1-x$, which is compatible with the standard negator defined for fuzzy logic. 

The negation doesn't change the uncertainty degree of the element, rather it can be viewed as a rotation of the fuzzy number in $\mathscr{T}$ with respect to the line $x = 0.5$. 

For example,

$\neg TFN(0.2,0.6,0.7) = TFN(0.3,0.4,0.8)$ and $\neg[TrFN(-2,0.3,0.9,3)] = [TrFN(-2,0.1,0.7,3)]$.

\subsubsection{Negation-as-failure($not$):}

Negation-as-failure($not$) is crucial to capture the nonmonotonicity of answer set programming. The significance of $not$ is that, it enables syntactical representation of incompleteness of knowledge in logic programs.

For a proposition $p$, $not \ p$ is to be true if nothing is known about $p$. It is notable that unlike $\neg p$, the truth of $not \ p$ doesn't require evidential refutation of $p$, rather we can perform reasoning even if the acquired knowledge about $p$ is incomplete. 

When nothing is known about $p$, the epistemic state $IFN(0,1)$ (or $TrFN(0,0,1,1)$) is assigned to $p$, which has uncertainty degree of $1$. For that, $not \ p$ is True, i.e. assigned with $IFN(1,1)$(or $TrFN(1,1,1,1)$). Hence, $not \ IFN(0,1) = not \ TrFN(0,0,1,1) = IFN(1,1)$. When some knowledge about $p$ is available, the epistemic state of $not \ p$ depends upon the degree of truth and uncertainty degree of $p$. Thus, the truth assignment of $not \ p$, is a \textit{meta-level} assertion, depending on the epistemic state of $p$, which is already determined. Thus, there is no inherent uncertainty in the epistemic state of $not \ p$ and because of this, the epistemic state of $not \ p$ would be an exact interval of the form $IFN(x,x)$, for $x \in [0,1]$. The negation-as-failure can be defined as:

$not \ IFN(a,b) = IFN(1-a,1-a)$.

$not \ \Box(a,b,c,d) = IFN(1-b,1-b) = not \ \Delta(a,b,c)$.

\subsubsection{Conjunction and Disjunction:}

T-representable product t-norm is used as conjunctor here. For two $IFN$s their product can be defined using interval algebra \cite{gao2009multiplication} as follows:

\begin{center}

$[a_1,a_2] \odot [b_1,b_2] = [min(a_1b_1, a_1b_2, a_2b_1, a_2b_2), max(a_1b_1, a_1b_2, a_2b_1, a_2b_2)]$

\end{center}

When $a_1,a_2,b_1,b_2$ are positive real $\in \Re^+$ then $[a_1,a_2] \odot [b_1,b_2] = [a_1a_2, b_1b_2]$.

The product t-norm of two restricted elements of $\mathscr{T}$ is defined from the standard approximated product of two fuzzy numbers \cite{giachetti1997analysis,dubois1993fuzzy,kaufmann1988fuzzy}.

$\bullet  \ IFN(a_1,b_1) \wedge IFN(a_2,b_2) = [a_1a_2,b_1b_2]$;

$\bullet  \ TFN(a_1,b_1,c_1) \wedge TFN(a_2,b_2,c_2) = TFN(a_1a_2, b_1b_2, c_1c_2)$;

$\bullet  \ TrFN(a_1,b_1,c_1,d_1) \wedge TrFN(a_2,b_2,c_2,d_2) = TrFN(a_1a_2, b_1b_2, c_1c_2, d_1d_2)$.

In case of semi-restricted fuzzy numbers the tnorms will be:

$\bullet \ [TFN(a_1,b_1,c_1)] \wedge [TFN(a_2,b_2,c_2)] = [TFN(min(a_1a_2, a_1c_2, c_1a_2, c_1c_2), b_1b_2,$ 

$max(a_1a_2, a_1c_2, c_1a_2, c_1c_2))]$.

$\bullet \ [TrFN(a_1,b_1,c_1,d_1)] \wedge [TrFN(a_2,b_2,c_2,d_2)] = [TrFN(min(a_1a_2, a_1d_2, d_1a_2, d_1d_2),$

$ min(b_1b_2, b_1c_2, c_1b_2, c_1c_2), max(b_1b_2, b_1c_2, c_1b_2, c_1c_2), max(a_1a_2, a_1d_2, d_1a_2, d_1d_2))]$.

The disjunction of $\mathscr{P}_1, \mathscr{P}_2 \in \mathscr{T}$ can be obtained from the standard negator($\neg$) and $\wedge$ by means of De Morgan's Law as follows:

$\mathscr{P}_1 \vee \mathscr{P}_2 = \neg [(\neg \mathscr{P}_1) \wedge (\neg \mathscr{P}_2)]$.

\subsubsection{Knowledge aggregation operator ($\otimes_k$)}

Apart from the aforementioned connectives, another connective is introduced for non-monotonic reasoning, which is the knowledge aggregation operator $\otimes_k$. For two elements $\mathscr{P}_1, \mathscr{P}_2 \in \mathscr{T}$, $\otimes_k$ is defined as follows:

$\mathscr{P}_1 \otimes_k \mathscr{P}_2 = \begin{cases} \mathscr{P}_1 & k_{\mathscr{P}_1} \geq k_{\mathscr{P}_2} \\  \mathscr{P}_2 & k_{\mathscr{P}_1} \leq k_{\mathscr{P}_2} \end{cases}$

Thus, $\otimes_k$ chooses which among the two truth assignments is more certain.
\subsection{Syntax:}

The language consists of infinitely many variables, finitely many constants and predicate symbols and no function symbol. For a predicate symbol $p$ of arity $n$, $p(t_1, t_2,...,t_n)$ is an atom, where $t_1,..t_n$ are variables or constants or an element of $\mathscr{T}$. A grounded atom contains no variable. A literal is a positive atom or its negation. For a literal $l$, $not \ l$ is a naf-literal.

An UnASP program consists of weighted rules of the form:

\begin{center}

$r:a \stackrel{\alpha_r}{\longleftarrow} b_1 \wedge ... \wedge b_k \wedge not \ b_{k+1} \wedge ... \wedge not \ b_n$

\end{center}

where, $\alpha_r \in \mathscr{T}$ is the weight of the rule, which denotes the epistemic state of the consequent or head ($a$) of the rule, when the antecedent or body ($b_1 \wedge ... \wedge b_k \wedge not \ b_{k+1} \wedge ... \wedge not \ b_n$) of the rule is true, i.e., has the truth status $IFN(1,1)$.

$a,b_1,...,b_n$ are positive or negative literals or elements of $\mathscr{T}$. For simplicity the body of the rule will be denoted by ',' separated literals instead of using the $\wedge$ symbols. A rule is said to be a \textbf{fact} if $b_i, 1 \leq i \leq n$ are elements of $\mathscr{T}$.

The rule weight $\alpha_r$ may denote the inherent uncertainty of the rule, or the degree of reliability or priority of the source of the rule. Even $\alpha_r$ can be used to denote that the rule '$r$' is a disposition, i.e. a proposition with exceptions.

\subsection{Declarative Semantics:}

The semantics is similar to the UnASP framework, proposed  in \cite{paul2019unified}; hence here it is specified briefly.

$\mathscr{L}$ be the set of literals (excluding naf-literals). An \textbf{interpretation}, $I$, is a set $\{a:\mathscr{P}_a|\mathscr{P}_a \in \mathscr{T}\}$, which specifies the epistemic states of the literals in the program.

\textbf{Definition 5:} An interpretation $I$ is inconsistent if there exists an atom $a$, such that, $a:\mathscr{P}_a \in I$ and $\neg a: \mathscr{P}_{\neg a} \in I$ and $k_{\mathscr{P}_a} = k_{\mathscr{P}_{\neg a}}$ but $t_{\mathscr{P}_a} \neq 1-t_{\mathscr{P}_{\neg a}}$.

In other words, an inconsistent interpretation assigns contradictory truth status to two complemented literals with same confidence.

The set of interpretations can be ordered with respect to the uncertainty degree by means of the knowledge ordering ($\leq_{k_p}$). For two interpretations $I$ and $I^*$, $I \leq_k I*$ iff $\forall a \in \mathscr{L}, I(a) \leq_{k_p} I^*(a)$. An interpretation $I_k$ is the \textbf{k-minimal} interpretation of a set of interpretations $\Gamma$, iff for no interpretation $I^* \in \Gamma$; $I^* \leq_{k_p} I_k$. If for any $\Gamma$, $I_k$ is unique then it is \textit{k-least}.

\textbf{Definition 6:} An interpretation $I$ \textit{satisfies} a rule $r$ if for every ground instance of $r$ of the form $r_g:head \stackrel{\alpha_r}{\longleftarrow} body$, $I(head) = (I(body)\wedge \alpha_r)$ or $I(head) >_{k_p} (I(body) \wedge \alpha_r)$ or $I(head) >_{t_P} (I(body) \wedge \alpha_r)$. $I$ is said to be a \textit{model} of a program $P$, if $I$ satisfies every rule of $P$.

\textbf{Definition 7:} A model of a program $P$, $I_m$, is said to be \textbf{supported} iff:

 1. For every grounded rule $r_g: a \stackrel{\alpha_r}{\leftarrow} b$, such that $a$ doesn't occur in the head of any other rule, $I_m(a) = I_m(b)$.

 2. For grounded rules $\{ a \stackrel{\alpha_1}{\leftarrow} b_1, a$     $\stackrel{\alpha_2}{\leftarrow} b_2,..,a \stackrel{\alpha_n}{\leftarrow} b_n\} \ \in P$ having same head $a$, $I_m(a) = (I_m(b_1) \wedge \alpha_1) \vee ... \vee (I_m(b_n) \wedge \alpha_n)$.

 3. For literal $l \in \mathscr{L}$, and grounded rules $r_l: \ l \longleftarrow b_l$, and $\ r_{\neg l}: \neg l \longleftarrow \ b_{\neg l}$, in $P$, $I_m(l) = I_m(b_l) \otimes_K \neg I_m(b_{\neg l})$ and $I_m(b_l) \otimes_k \neg I_m(b_{\neg l})$ exists in $\mathscr{T}$. 

The first condition of supportedness guarantees that the inference drawn by a rule is no more certain and no more true than the degree permitted by the rule body and rule weight. The second condition specifies the optimistic way of combining truth assertions for an atom coming from more than one rule. The third condition captures the essence of nonmonotonicity of reasoning. For an atom $a$, rules with $a$ in the head are treated as evidence in favour of $a$ and rules with $\neg a$ in the head stands for evidence against $a$. In such a scenario, the conclusion having more certainty or reliability is taken as the final truth status of $a$.

\textbf{Definition 8:} The \textbf{reduct} of a program $P$ with respect to an interpretation $I$ is defined as:

$P^I = \{r_I:a \stackrel{\alpha_r}{\longleftarrow} b_1 \wedge ... \wedge b_k \wedge not \ I(b_{k+1}) \wedge ... \wedge not \ I(b_n) \ | \ r \in P\}$.

$P^I$ doesn't contain any naf-literal in any rule. For a positive program $P$ (with no rules ontaining $not$), $P^I = P.$

\textbf{Definition 9:} For any UnASP program $P$, an interpretation $I$ is an \textbf{answer set} if $I$ is an k-minimal supported model of $P^I$. For a positive program the k-minimal model is unique.

\textbf{Example 8:} The motivating example described in \cite{bauters2010towards} is considered here. 

$P=\{ r1: tumor \xleftarrow{[TFN(0.4,0.4,1.5)]} cin_{on} \wedge tsg_{off}$ 

$r2: tumor \xleftarrow{TFN(0.1,0.1,0.5)} tsg_{off}$

$r3: tsg_{off} \xleftarrow{IFN(0.6,1)} cin_{on} \}$

These rules describe the same information described there by means of 9 rules in a lot more brief and intuitive way. When $tsg_{off}$ and $cin_{on}$ both holds rule $r1$ infers $tumor:[TFN(0.4,0.4,1.5)]$. This essentially a similar truth assertion as derived in \cite{bauters2010towards} as $\{tumor^{0.8:0.4}, tumor^{0.6:0.6}, tumor^{0.4:0.8}, tumor^{0.2:1}\}$. Rule $r3$ signifies that when $cin_{on}$ holds there is a chance that $tsg_{off}$ holds; the underlying uncertainty is depicted by the interval $[0.6,1]$. This example illustrates the effectiveness of the developed framework.

\section{Conclusion:}

This paper explores the feasibility of considering fuzzy numbers as truth values of propositions. Unlike a uniform interval, fuzzy numbers defined over the interval $[0,1]$ can be viewed as an interval with a membership distribution defined over it. Here, mainly triangular and trapezoidal membership distributions are considered. Using them as the set of truth values greatly enhance the expressive power of a logical framework. The truth values or epistemic states are ordered with respect to degree of truth and degree of certainty by means of a preorder-based algebraic structure. The truth and knowledge orderings defined here are intuitive and also enable performing nonmonotonic reasoning using uncertain and imprecise information. To demonstrate the effectiveness of the modified truth value space an answer set programming framework is developed over it. This type of framework can be utilized in decision support systems or as the logical system underlying a semantic web to represent the underlying uncertainty and vagueness.

Using the fundamental idea behind defining truth and knowledge ordering, other membership distributions, like \textit{sigmoid}, \textit{gaussian} defined over the interval $[0,1]$, may be fitted in $\mathscr{T}$, if necessary. Even \textit{bimodal} or \textit{multi-modal} distributions can be used and ordered using $\leq_{t_p}$ and $\leq_{k_p}$ as defined here. Thus more accurate representation of various real life situations is attainable.

\textbf{Acknowledgment:} The first author acknowledges the scholarship obtained from Department of Science and Technology, Government of India, in the form of INSPIRE Fellowship.

\bibliographystyle{spmpsci}      
\bibliography{biblist3}

\end{document}